\documentclass{article}

\usepackage{microtype}
\usepackage{graphicx}
\usepackage{subcaption}
\usepackage{booktabs}
\usepackage{hyperref}
\usepackage{textcomp}

\usepackage[accepted]{icml2026}
\usepackage{stfloats}
\usepackage{amsmath,amssymb,mathtools,amsthm}
\usepackage{enumitem}
\usepackage{tikz}
\usetikzlibrary{positioning,arrows.meta,calc,decorations.pathreplacing}
\usepackage[capitalize,noabbrev]{cleveref}
\theoremstyle{plain}
\newtheorem{theorem}{Theorem}[section]
\newtheorem{proposition}[theorem]{Proposition}

\theoremstyle{definition}
\newtheorem{definition}[theorem]{Definition}
\theoremstyle{remark}

\newcommand{\loss}{\mathcal{L}}
\newcommand{\R}{\mathbb{R}}

\newcommand{\Pos}{\mathcal{P}}
\newcommand{\Neg}{\mathcal{N}}

\DeclareMathOperator*{\argmax}{arg\,max}

\icmltitlerunning{Cascading Circle Loss for Graded Legal Retrieval}

\begin{document}

\twocolumn[
  \icmltitle{Cascading Circle Loss: Order-Preserving\\
  Contrastive Learning for Graded Legal Retrieval}

\begin{icmlauthorlist}
    \icmlauthor{Rayane Kimouche}{ups}
    \icmlauthor{Ahmed Touila}{ups}
\end{icmlauthorlist}

  \icmlaffiliation{ups}{
    Dilitrust, Paris, France
  }

  \icmlcorrespondingauthor{Rayane Kimouche}{rayane.kimouche@dilitrust.com}
  \icmlcorrespondingauthor{Ahmed Touila}{ahmed.touila@dilitrust.com}

  \icmlkeywords{Contrastive Learning, Legal Retrieval, Graded Relevance, Metric Learning, Circle Loss}

  \vskip 0.3in
]
\printAffiliationsAndNotice{}

\begin{abstract}
This paper introduces \textbf{Cascading Circle Loss (CasCL)}, a contrastive learning framework for legal retrieval tasks with graded relevance. CasCL combines an \textbf{InfoNCE} objective for positive--negative separation with a cascade of ranking boundaries that model ordered relevance levels between queries and legal clauses. The framework also includes a \textbf{negative-aware self-paced optimization} mechanism and a \textbf{relevance gain weighting} strategy to improve ranking quality. Experiments on legal passage retrieval benchmarks show that CasCL consistently outperforms existing methods across both binary retrieval and graded ranking metrics. The implementation is available at
\url{https://github.com/RayaneKimo/Cascading-Circle-Loss}.
\end{abstract}

\section{Introduction}
\label{sec:intro}

Legal professionals rely on retrieval systems to surface the most relevant paragraphs and extracted legal information for a given query ~\cite{locke2022caselaw, ferraro2022legal}. Unlike generic information retrieval, where a single best match is often sufficient, legal retrieval requires a ranked slate of outputs: paragraphs that explicitly contain target labels or directly relevant extracted entities must be ranked above paragraphs that are only loosely related through shared wording or contextual similarity, which in turn must precede minimally related passages.
This graded notion of relevance is fundamental in legal applications ~\cite{saracevic2007relevance, jarvelin2002cumulated}, as incorrect ordering can lead to incorrect information extraction, poor downstream reasoning, and degraded decision-making quality~\cite{park2025legal, barnert2025reliable}.

 \paragraph{The graded legal relevance problem.}
   Unlike web search, where a single best result often suffices, legal
   queries are rarely answered by a single paragraph in isolation. A
   query such as \emph{what are the penalties for late delivery under
   this agreement ?}\ sits inside a dense neighbourhood of related
   material~\cite{vanopijnen2017relevance}: the clause defining timely
   delivery, the liability cap bounding the penalty, the force majeure
   provision that may suspend it, and the governing law clause that
   determines interpretation. Each is topically related, yet only one
   directly answers the query; the rest occupy intermediate positions on
   a relevance spectrum.

 Legal corpora exhibit three structural properties that make fine-grained ranking essential. First, \textbf{high lexical overlap}: statutes and contractual clauses use highly standardized language, so neighbouring paragraphs often share near identical phrasing while encoding distinct legal conditions ~\cite{tsarapatsanis2025lexical, choi2025acord}, binary supervision cannot distinguish them. Second, \textbf{hierarchical relevance}: a legal topic (e.g liability, confidentiality) is rarely confined to a single paragraph but distributed across definitions, exceptions, and remedies, inducing a continuous relevance spectrum that discrete labels fail to capture ~\cite{vanopijnen2017relevance, ma2024legal} . Third,
 \textbf{sensitivity to misranking}: when the top of the ranked slate mixes directly applicable clauses with merely related ones, ordering errors propagate to downstream tasks,  extraction pipelines pull wrong
 spans, summarizers conflate distinct legal meanings, and reasoning systems inherit ranking errors as factual errors ~\cite{park2025legal, barnert2025reliable}.

A loss function suited to this setting therefore needs four properties simultaneously:

 \begin{enumerate}[nosep,leftmargin=*]
   \item \textbf{Robust positive–negative separation}, so the retriever still excels at the binary task that existing systems are evaluated on;
   \item \textbf{Intra-positive ordering:} among relevant passages, higher relevance must yield higher similarity, with gaps reflecting the relevance difference.
 \item \textbf{Top-of-ranking focus:} not all ordering errors are
     equally costly~\cite{jarvelin2002cumulated}; swapping positions 1
     and 2 matters far more than swapping 10 and 11. The loss should
     concentrate gradient on the top of the ranking.
   \item \textbf{Robustness to ambiguous relevance:} in large batch training with graded labels, lower ranked items tend to carry more ambiguous relevance scores ~\cite{ma2023relevance}, a passage scored at r = 0.15 or r = 0.10 may reflect annotator (or LLM) noise ~\cite{zheng2023judging} rather than a genuine relevance difference. The loss should down weight ordering constraints where label uncertainty is highest.
 \end{enumerate}

\paragraph{Existing approaches fall short.}
 
   No existing contrastive loss satisfies all four properties simultaneously.
   Binary losses such as InfoNCE~\cite{oord2018representation} address property~1
   but ignore intra-positive ordering entirely~\cite{wullach2026graded,esfandiarpoor2025beyond}.
   Self-paced losses like Circle Loss~\cite{sun2020circle} lack inter-positive structure.
   Chain-based approaches such as Ladder Loss~\cite{zhou2019ladder} enforce ordering
   but with fixed margins and no self-pacing.
   Recent graded extensions, CLOC~\cite{pitawela2025cloc},
   Mine \& Refine~\cite{xi2026mine},
   and soft-target variants~\cite{zhu2025gcl,denize2022sce}, either
   require multi-stage pipelines, discretise the relevance spectrum,
   or provide no geometric ordering guarantee.
   We position CasCL against each of these methods in detail in \S\ref{sec:related}.

   \paragraph{Our approach: modular decomposition.}
   We observe that positive--negative separation (property~1) and
   intra-positive ordering (property~2) are \emph{structurally
   independent}: the first is a binary discrimination problem, the
   second a fine-grained relative comparison within the positive set.
   CasCL decouples them: InfoNCE sees only the best positive and all
   negatives, establishing the absolute scale; a cascade of self-paced
   circular boundaries orders the graded positives, with each item
   invisible to the other term. The design rationale is detailed in
   \S\ref{sec:why-modular}.

\paragraph{Contributions.}
To address these limitations, CasCL introduces the following contributions:
\begin{enumerate}[nosep,leftmargin=*]
\item \textbf{CasCL (\S\ref{sec:method}):} A modular loss combining
     InfoNCE for binary separation and a cascade of self-paced circular
     boundaries for intra-positive ordering, extending Circle Loss's
     self-paced mechanism to \emph{positive--positive} pairs for the
     first time.
 \item \textbf{Relevance gain weighting (\S\ref{sec:relgain}):} Each
     cascade link is weighted by $w_j = 2^{r_j}\!-\!1$, concentrating
     gradient on high relevance ordering decisions and suppressing noise
     from ambiguous low relevance labels.

\item \textbf{Negative-aware two-regime self-pacing (\S\ref{sec:tworegime}):}
         A dual regime self-paced weight that prevents graded positives
         from being stranded below negatives with zero gradient,
         resolving the MAP/nDCG tension.
        
  \item \textbf{Order guarantee (\S\ref{sec:theory}):} Proof that adjacent link satisfaction implies full ranking with cumulative margins $\delta_0 \cdot |r_i - r_j|$ at $O(K)$ cost.
  \item \textbf{Gradient analysis (\S\ref{sec:gradients}):} Each item receives gradients from at most two neighbours, creating a spring system with automatic focus on the most violated link.
\end{enumerate}

\section{Related Work}
\label{sec:related}

   \paragraph{Standard InfoNCE.~\cite{oord2018representation}} is the $K\!=\!1$ special case of CasCL: it provides robust positive--negative separation (property~1) but treats all non positive items identically, ignoring intra positive ordering entirely
(property~2)~\cite{wullach2026graded,esfandiarpoor2025beyond}. 
 \paragraph{Soft Target Extensions: GCL~\cite{zhu2025gcl} and SCE~\cite{denize2022sce}} accommodate graded relevance inside the softmax, yet the partition function still pulls all
 positives toward equally high similarity, conflating ``most relevant'' with ``somewhat relevant,'' and provides no geometric structure or explicit ordering guarantee; CasCL keeps graded positives outside the InfoNCE term entirely, delegating
 their ordering to a dedicated cascade.
   \paragraph{Circle Loss.~\cite{sun2020circle} }introduced self-paced weights that vanish at target similarity, providing an efficient training mechanism that concentrates gradient on the most misplaced pairs, but it operates only on independent
 positive--negative pairs with no inter-positive structure (property~2) and no notion of ranking position (properties~3 and~4); CasCL adapts Circle Loss’s self-paced mechanism to inter-positive ordering by chaining circular boundaries between adjacent items in the ranking rather than class pairs.
   \paragraph{Ladder Loss.~\cite{zhou2019ladder}} also chains ordering constraints across ranked items (property~2), but uses fixed discrete margins with no self-pacing: every link contributes gradient regardless of whether it is already satisfied, wasting
 capacity on satisfied constraints and amplifying noise from ambiguous labels (violating properties~3 and~4); CasCL replaces fixed margins with continuous $\delta_j \propto \Delta r$ and adds self-paced weights that shut off satisfied links,
 focusing capacity on the violations that remain.
   \paragraph{CLOC.~\cite{pitawela2025cloc}} learns cumulative margins for ordinal ranks but requires two-phase training (joint optimisation followed by margin freezing) to prevent collapse; CasCL derives margins analytically from $\Delta r$ with no
 learning phase and no staged schedule.
   \paragraph{Mine \& Refine.~\cite{xi2026mine}} extends Circle Loss to three discrete relevance levels, each with a manually tuned boundary, but three structural gaps remain: \emph{(i)}~two items labelled ``moderate'' ($r\!=\!0.3$ and $r\!=\!0.8$) receive
 identical treatment, whereas CasCL assigns each item a unique margin proportional to its exact $\Delta r$, \emph{(ii)}~nothing prevents a ``moderate'' item from outscoring a ``relevant'' one, since per-level circles are satisfied in isolation
 with no cross-level ordering guarantee (property~2), whereas CasCL's cascade chain explicitly links all levels, making such inversions impossible (Theorem~\ref{thm:order}) and \emph{(iii)}~M\&R requires a two-stage pipeline (SupCon then
 multi-class Circle Loss) with eight manually tuned boundary parameters,  whereas CasCL is end-to-end with a single balance weight~$\mu$
   and one base margin~$\delta_0$, all per-item margins derived
   automatically.

\section{Method: Cascading Circle Loss}
\label{sec:method}

\subsection{Why Modular Decomposition}
\label{sec:why-modular}

\begin{figure}[ht]
  \vskip 0.2in
  \begin{center}
\centerline{\includegraphics[width=\columnwidth]{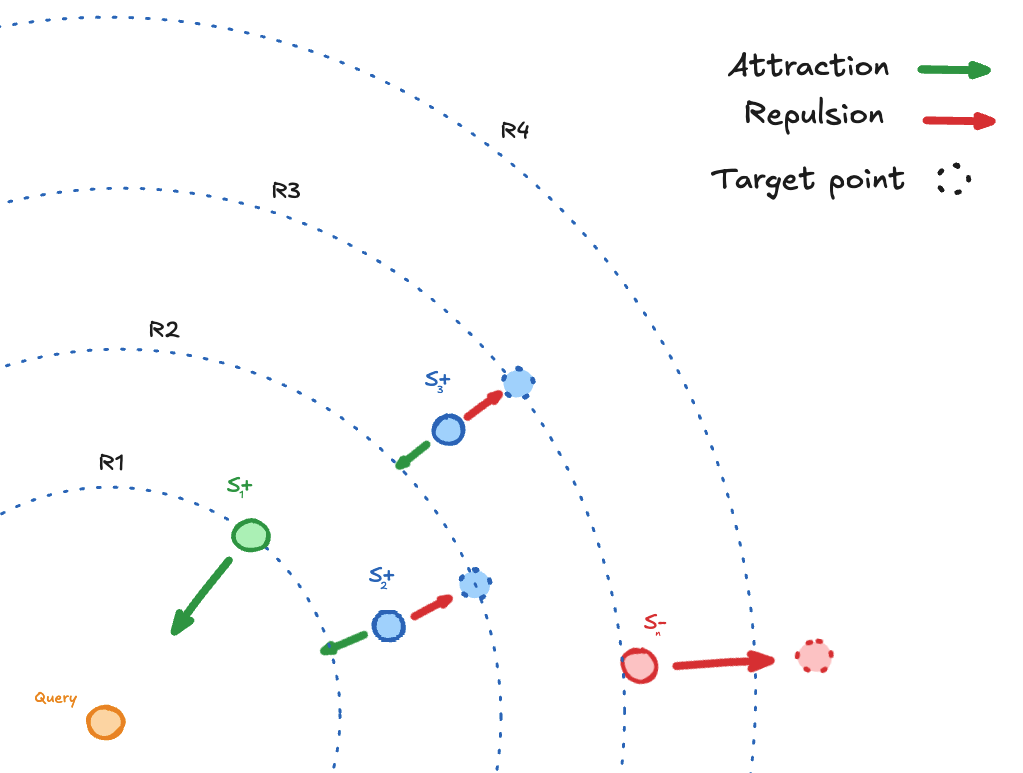}}
    \caption{
    Embedding space geometry induced by CasCL. Graded positives ($s_1^+$, $s_2^+$, $s_3^+$) are arranged into concentric relevance rings ($R_1$–$R_4$) around the query, with higher-relevance items occupying inner rings. Green arrows
 indicate the cascade's attraction force pulling each positive toward its relevance proportional equilibrium; red arrows indicate the repulsion force pushing items that are too close outward toward their target ring. The negative ($s_n^-$) is
 repulsed beyond the outermost ring. Dashed circles represent the target positions derived from the self-paced weights, and the margin between consecutive rings scales proportionally with the relevance gap $\delta_j = \delta_0 \cdot \Delta r$.
    }
    \label{spacegeometry}
  \end{center}
\end{figure}
One might try a single ``graded InfoNCE'' with soft targets $p_j \propto r_j$.
This creates a tension: should a positive with a low relevancy $(r = 0.2)$ be pulled toward a higher similarity $s \approx 1$ (as the softmax pressure does) or left at moderate similarity (as its low relevance suggests) ?
Soft target losses implicitly pull \emph{all} positives toward high similarity, they cannot express ``this item is positive but should not be too close.'' %
Placing a graded positive in the InfoNCE numerator pulls it too close; placing it in the denominator pushes it too far; neither treatment is correct.

In this work, we extend Circle Loss to graded relevance by adjusting the optimal target $O_p(r)$.
But this requires choosing a target function $s^*(r)$ mapping relevance to similarity, and any such function embeds assumptions about the ``right'' similarity for each relevance level.

Our solution is to embed the relevance score
 directly into the self-paced weights of the cascade:
\begin{itemize}[nosep]
  \item InfoNCE handles the \emph{best positive}, the only one that unambiguously should be at $s \approx 1$, using the standard binary formulation.
  \item The cascade uses each item's relevance $r_j$ as a natural equilibrium point in its self-paced weight $\alpha_j = [r_j - s_j]_+$, while the chain structure and proportional margins enforce the ordering between these
 positions.
\end{itemize}

\subsection{The CasCL Loss}

\begin{figure}[ht]
  \vskip 0.2in
  \begin{center}
    \centerline{\includegraphics[width=\columnwidth]{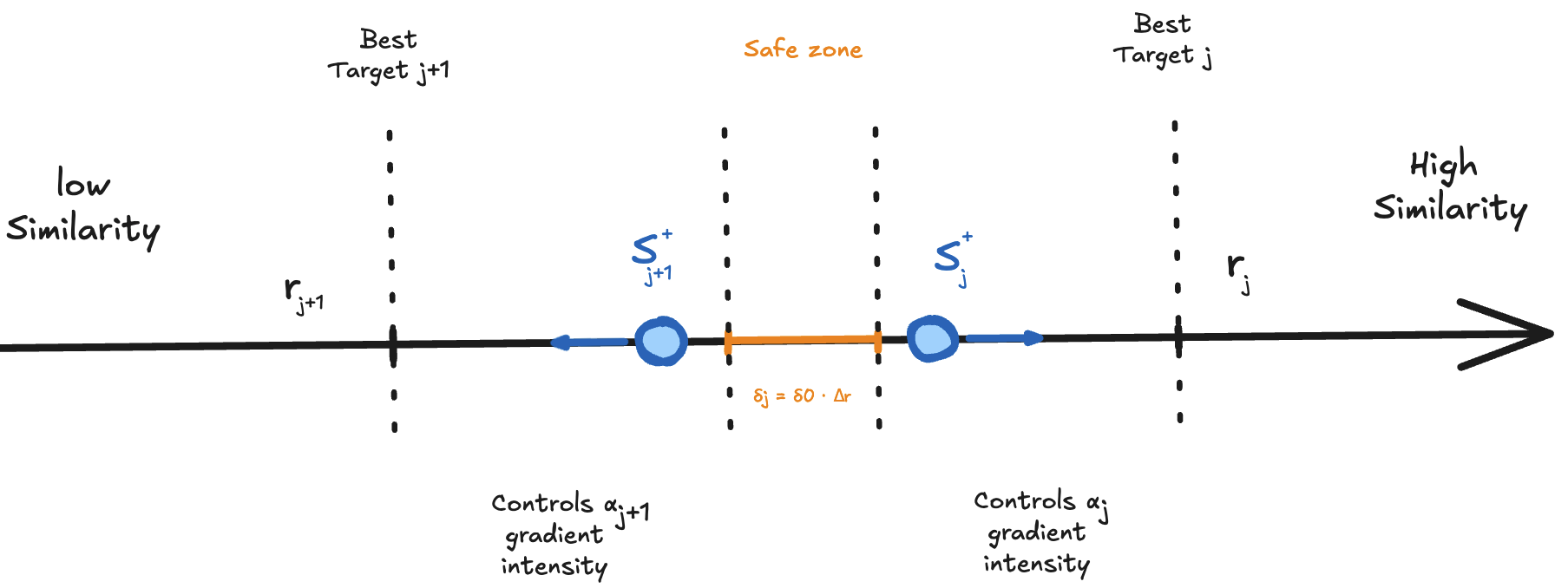}}
    \caption{
    Anatomy of a single cascade link between two adjacent graded positives $s_j^+$ and $s_{j+1}^+$ on the similarity axis. Each item is pulled toward its relevance-derived equilibrium ($r_j$ and $r_{j+1}$), which controls the self-paced
 gradient intensity ($\alpha_j$ and $\alpha_{j+1}$). The orange segment between the two items represents the safe zone, the minimum margin $\delta_j = \delta_0 \cdot \Delta r$ that the cascade enforces between them. The higher-relevance item
 $s_j^+$ is pushed right toward $r_j$; the lower-relevance item $s_{j+1}^+$ is pushed left toward $r_{j+1}$. Gradient magnitude decays as each item approaches its target, shutting off entirely at the equilibrium point.
    }
    \label{single-cascade-link}
  \end{center}
\end{figure}

The CasCL loss addresses both positive–negative separation and intra-positive ordering in a single modular objective
 where each component operates on a distinct subset of items with no overlap in function, see the global desired behavior of our loss in Figure~\ref{spacegeometry}. 
\begin{definition}[Cascading Circle Loss]
\label{def:cascl}
\begin{equation}
\boxed{
  \loss_{\text{CasCL}} = \loss_{\text{InfoNCE}}(s_1; \Neg, \tau) \;+\; \mu \cdot \loss_{\text{cascade}}
}
\label{eq:cascl-full}
\end{equation}
where the InfoNCE term is:
\begin{equation}
  \loss_{\text{InfoNCE}} = -\log \frac{e^{s_1/\tau}}{e^{s_1/\tau} + \sum_{k \in \Neg} e^{s_k/\tau}}
  \label{eq:infonce-term}
\end{equation}
and the cascade term is:
\begin{equation}
  \loss_{\text{cascade}} = \sum_{j=1}^{K} w_j \;\log\!\Bigl[1 + e^{\,\gamma(\alpha_{j+1} s_{j+1} - \alpha_j s_j + \delta_j)}\Bigr]
  \label{eq:cascade-term}
\end{equation}
with relevance gain weight $w_j = 2^{r_j}\!-\!1$ (\S\ref{sec:relgain}).
\end{definition}

The two terms operate on \textbf{disjoint item sets} with \textbf{complementary roles}:

\paragraph{Term 1: InfoNCE (Scale Setting).}
Standard binary InfoNCE using $d_1$ (most relevant positive, $r_1 = \max_i r_i$) as the positive and \emph{all} negatives in the denominator.
This establishes the absolute scale : $s_1$ is pushed high, all $s_k$ for $k \in \Neg$ are pushed low.
Graded positives $d_2, \ldots, d_K$ do \emph{not} appear in this term, they are invisible to InfoNCE.

\paragraph{Term 2: Cascade (Order Preserving).}
\label{sec:cascade}
A chain of $K$ self-paced circular links along the sorted sequence $d_1 \succ d_2 \succ \ldots \succ d_K \succ d_{\text{neg}}$, where $d_{\text{neg}} = \argmax_{k \in \Neg} s_k$ is the hardest negative.

   The cascade's last link uses only the hardest negative
   $d_{\text{neg}}$; all other negatives participate exclusively through
   InfoNCE. This suffices because if $d_K$ beats the hardest negative, it
   beats all negatives by definition, and InfoNCE pushes all negatives
   down globally.

\begin{figure}[ht]
  \vskip 0.2in
  \begin{center}
    \centerline{\includegraphics[width=\columnwidth]{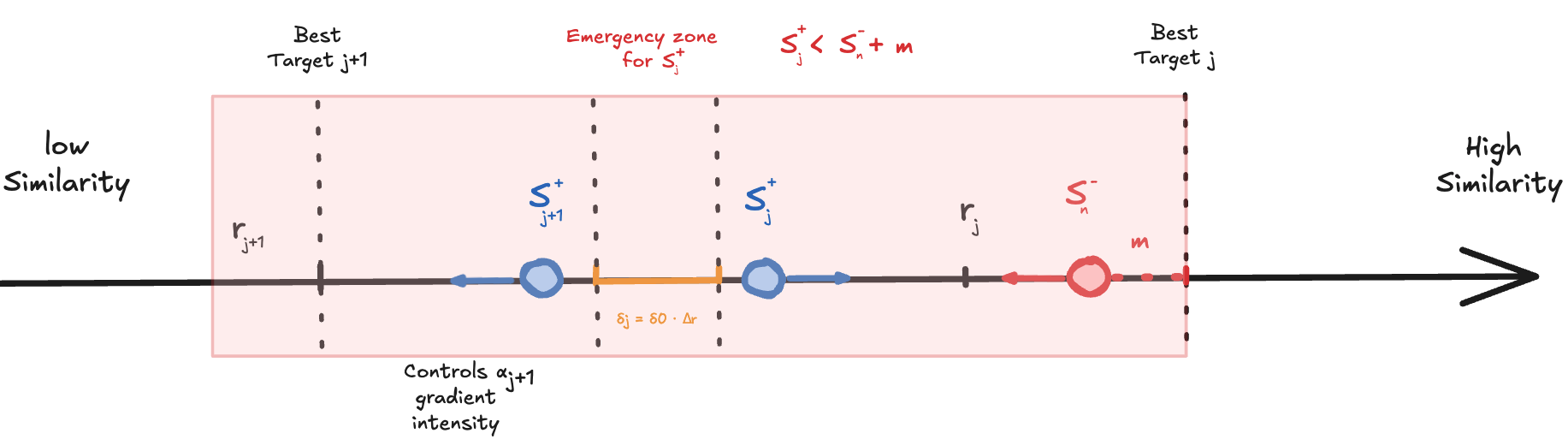}}
    \caption{
       Emergency mode for a single cascade link. Item $s_j^+$ (higher
       relevance) is stranded below the emergency boundary
       ($s_{\text{neg}} + m_{\text{safe}}$, red dashed line), inside the
       emergency zone (red shaded region). In standard self-pacing,
       $\alpha_j$ would vanish at $r_j$, leaving $s_j^+$ stranded below
       the hardest negative with no corrective gradient. The two-regime
       mechanism activates the emergency term:
       $\alpha_j = [s_{\text{neg}} + m_{\text{safe}} - s_j]_{+}$, which
       keeps the self-paced weight large and produces a strong upward
       pull on the positive to restore it above the negative boundary,
       while the cascade's last link simultaneously pushes
       $s_{\text{neg}}$ downward. Once $s_j^+$ clears
       $s_{\text{neg}} + m_{\text{safe}}$, the emergency term deactivates
       and standard refinement toward $r_j$ resumes. The orange segment
       represents the proportional margin
       $\delta_j = \delta_0 \cdot \Delta r$ enforced between the two
       items.
       }
    \label{emergency-example}
  \end{center}
\end{figure}

For link $j$ connecting $(d_j, d_{j+1})$, Figures~\ref{single-cascade-link} and ~\ref{emergency-example}, the per-link components are:
\begin{itemize}[nosep,leftmargin=*]
  \item \textbf{Upper weight} (two-regime, \S\ref{sec:tworegime}):
  \begin{equation}
    \alpha_j = \max\!\bigl([r_j\!-\!s_j]_+,\;[s_{\mathrm{neg}}\!+\!m_{\mathrm{safe}}\!-\!s_j]_+\bigr)
    \label{eq:alpha-up}
  \end{equation}
  \item \textbf{Lower weight} (absolute):
        $\alpha_{j+1} = [s_{j+1} - r_{j+1}]_+$ \label{eq:alpha-down}
  \item \textbf{Margin} (proportional):
        $\delta_j = \delta_0 \cdot (r_j - r_{j+1})$ \label{eq:delta}
\end{itemize}
with $r_{K+1} = 0$ for the last link and $s_{\mathrm{neg}} = \max_{k \in \Neg} s_k$.

 When $r_j = r_{j+1}$, $\delta_j = 0$ and the link allows
   $s_j = s_{j+1}$, requiring no special handling for ties.

   \noindent
      The original Circle Loss~\citep{sun2020circle} uses self-paced
   weights with \emph{constant} optimal targets ($O_p = 1\!+\!m$ for
   all positives, $O_n = -m$ for all negatives), designed for binary
   labels where every positive shares the same target similarity.
   CasCL replaces these constants with relevance derived targets and
   governs each cascade link with \textbf{two self-paced weights}. The
   upper weight $\alpha_j = [r_j - s_j]_+$ drives the higher relevance
   endpoint toward the query, vanishing as $s_j$ approaches its target
   $r_j$; exceptionally, the emergency term reactivates it when the item
   falls below the negative boundary (\S\ref{sec:tworegime}). The lower
   weight $\alpha_{j+1} = [s_{j+1} - r_{j+1}]_+$ drives the
   less relevant endpoint away from the query, vanishing as $s_{j+1}$
   settles at $r_{j+1}$. Because every graded positive appears as the
   \textbf{upper endpoint of one link and the lower endpoint of the
   next}, it receives two opposing control forces, one pulling it
   closer, one pushing it away, forming a \textbf{self-correcting
   spring system} that positions each item at its
   relevance proportional equilibrium. The margin $\delta_j$ scales with
   the relevance gap, giving similar items tiny margins and dissimilar
   items large ones.


\subsection{Negative Aware Two-Regime Self-Pacing}
\label{sec:tworegime}

The self-paced weights described above replace Circle Loss's constant
   targets with relevance proportional ones, positioning each graded
   positive at a similarity reflecting its relevance and shutting off
   once the target is reached. This is sufficient for
   \textbf{intra-positive ordering}, but it is \textbf{agnostic to where
   the negatives are}: if a graded positive with $r_j = 0.3$ drifts
   below the hardest negative during training, the upper weight
   $\alpha_j = [r_j - s_j]_+$ still vanishes at $s_j = r_j = 0.3$,
   declaring the item ``on target'' even though it is stranded below the
   decision boundary. The spring system self corrects ordering among
   positives, but has no force to rescue a positive that has fallen into
   negative territory.

\paragraph{Two-regime upper weight.}
We add an \emph{emergency} term that activates when items fall below the negative boundary:
\begin{equation}
  \alpha_j = \max\!\bigl(\underbrace{[r_j - s_j]_+}_{\text{refinement}},\;\;\underbrace{[s_{\mathrm{neg}} + m_{\mathrm{safe}} - s_j]_+}_{\text{emergency}}\bigr)
  \label{eq:two-regime}
\end{equation}
The weight is the maximum of two terms, yielding three regimes:

\begin{center}
\scriptsize
\setlength{\tabcolsep}{3pt}
\begin{tabular}{@{}lll@{}}
\toprule
Regime & Condition & Effect \\
\midrule
Emerg.   & $s_j\!<\!s_{\mathrm{neg}}\!+\!m$  & Push UP \\
Refine.  & $s_j\!>\!s_{\mathrm{neg}}\!+\!m$, $s_j\!<\!r_j$  & Self-paced \\
Off      & $s_j\!>\!r_j$, $s_j\!>\!s_{\mathrm{neg}}\!+\!m$  & $\alpha\!=\!0$ \\
\bottomrule
\end{tabular}
\end{center}

\paragraph{Why the lower weight stays absolute.}
The lower weight $\alpha_{j+1} = [s_{j+1} - r_{j+1}]_+$ remains
   purely relevance based: making it negative aware would create dead
   zones where both weights vanish despite an ordering violation
   ($s_{j+1} > s_j$ with no corrective gradient). The asymmetry is
   intentional, the upper weight serves dual duty (self-paced
   refinement toward $r_j$ in normal conditions, and boundary rescue
   when the item falls below negatives), while the lower weight focuses
   solely on pushing items that sit closer than their relevance warrants
   back toward their equilibrium.

   \subsection{Order Guarantee}
   \label{sec:order-guarantee}

   \begin{theorem}[Cascading Order Guarantee]
   \label{thm:order}
   If all $K$ cascade links have zero loss, then for any pair $(i, j)$
   with $r_i > r_j$:
   \begin{equation}
     s_i - s_j \geq \delta_0 \cdot (r_i - r_j)
   \end{equation}
   \end{theorem}

   \noindent The full ranking is guaranteed via transitivity at $O(K)$
   cost. The proof is in Appendix~\ref{app:proof}.

      \section{Gradient Analysis}
   \label{sec:gradients}

   Each item $d_j$ in the cascade receives gradients from at most two
   links (Proposition~\ref{prop:grad}):
   \begin{equation}
   \frac{\partial \mathcal{L}_{\text{cascade}}}{\partial s_j} = \underbrace{+w_{j-1} \gamma [s_j - r_j]_+ \sigma_{j-1}}_{G_j^{\downarrow}\text{ (link above, repulses)}} -\underbrace{w_j \gamma \alpha_j \sigma_j}_{G_j^{\uparrow}\text{ (link
 below, attracts)}}
     \label{eq:grad-cascade}
   \end{equation}
   where $\sigma_l$ is the sigmoid function and
   $w_l = 2^{r_l}\!-\!1$. Each graded positive undergoes two forces
   from its adjacent links: a \textbf{repulsion from above}, exerted by
   the more relevant neighbour to prevent the item from approaching the
   query beyond its relevance warrants, and an \textbf{attraction from
   below}, pulling it away from the less relevant neighbour to keep it
   above the next item in the ranking. Together, these two forces
   position each soft positive at the similarity that satisfies its
   relevance level. Graded positives receive \emph{zero} gradient from
   InfoNCE, only the cascade acts on them, and the two-regime
   $\alpha_j$ ensures that items stranded below the hardest negative
   receive emergency strength gradient regardless of their relevance
   target (Proposition~\ref{prop:focus}). The full derivation, gradient
   table, and self-paced shutoff analysis are in
   Appendix~\ref{app:gradient}.

\section{Experiments}
\label{sec:experiments}

\subsection{Setup}

\textbf{Corpus.}
   We evaluate on a proprietary multilingual legal clause retrieval dataset (DiliTrust internal data).
   The corpus comprises contractual clauses spanning six languages: French, Portuguese, English, Italian, Spanish, and German.
   This multilingual setting reflects the operational reality of cross border legal practice, where a single contract portfolio routinely contains clauses drafted in multiple jurisdictions and languages.

   \textbf{Query--passage construction.}
   Each instance consists of a legal query paired with one human validated positive passage (the directly answering clause, $r = 1.0$), a set of graded soft positives at intermediate relevance levels ($r \in \{0.2, 0.4, 0.6, 0.8\}$), and hard
 negatives ($r = 0$).
   Human annotators intervened exclusively at the positive selection stage: domain experts identified the single most relevant clause for each query.
   This design choice is deliberate, at scale, asking human annotators to consistently assign five-level graded relevance scores across 150K+ examples is neither feasible nor reliable, as inter-annotator agreement on fine-grained ordinal
 scales is notoriously low in legal corpora, where even experts disagree on the boundary between ``directly applicable'' and ``strongly related''.
   Instead, we anchor the supervision on the one judgment humans make reliably (binary: ``this clause answers the query'') and delegate the graded scoring to an automated pipeline with statistical safeguards described below.

   \textbf{Relevance scoring via LLM judge ensemble.}
   Graded relevance scores were assigned using an ensemble of three LLM judges (GPT 5.1, Qwen3.5-122B-A10B, Gemini 3.1 pro).
   Each model independently scored every candidate clause against the query, conditioned on the human selected positive passage as an anchor, following the prompt protocol described in the github.
   To mitigate individual model biases, a well documented concern in LLM-as-a-judge settings~\cite{zheng2023judging}, we adopted a \textbf{majority vote aggregation} strategy: the final relevance label for each clause is the score agreed upon by at least two of the three judges.
   This ensemble approach reduces the variance introduced by any single model's idiosyncratic scoring behaviour, yielding more stable and reproducible relevance labels than single judge alternatives.

   \textbf{Training and evaluation splits.}
   The training set comprises approximately \textbf{150K examples} after filtering; the test set contains \textbf{30K examples}.
   The raw pool prior to filtering was substantially larger, but we applied a label diversity constraint: each training example is retained only if its set of graded positives contains \textbf{at least three distinct relevance levels}
 (excluding the anchor positive at $r = 1.0$ and the hardest negative at $r = 0$).
   This filtering ensures that every training instance exercises the cascade's ordering mechanism across a meaningful portion of the relevance spectrum, preventing degenerate cases where the cascade chain collapses to a single link.

   \textbf{Negative mining strategy.}
   The dataset construction is tailored to the two term structure of CasCL\@.
   For the \textbf{InfoNCE term}, each query is paired with its human selected positive ($r = 1.0$) and a set of hard negatives ($r = 0$) that are topically unrelated to the query.
   For the \textbf{cascade term}, the chain is constructed from the positive (highest relevance), the soft positives (intermediate relevance levels), and the \textbf{hardest negatives}, specifically the negative passages with the highest
 cosine similarity to the query embedding.
   This design is critical: by placing the most confusable negatives at the bottom of the cascade chain, we ensure that satisfying the adjacent link constraints forces all soft positives above all negatives via transitivity
 (Theorem~\ref{thm:order}).
   A random negative with near zero similarity would trivially satisfy the last cascade link, providing no gradient signal for the ordering of soft positives relative to the decision boundary.

   \textbf{Label distribution.}
   Relevance scores are distributed across five discrete levels: $r \in
   \{0.2, 0.4, 0.6, 0.8, 1.0\}$, each reflecting the degree to which a
   passage addresses the legal query as assessed by the LLM judge
   ensemble. A score of $r = 1.0$ indicates a directly applicable clause
   (human validated), while $r = 0.2$ indicates a tangentially related
   passage sharing legal context but not substantively answering the
   query. The judges score along two axes, \textbf{concept overlap}, does
   the candidate share the same legal concept as the positive, e.g.\
   both are confidentiality clauses ? and \textbf{information type match}, does it contain the same type of key information, e.g.\ a date, a
   monetary amount ? which are combined into the final ordinal rank;
   the full prompt is provided in the repository. With this uniform
   spacing, adjacent levels share a constant gap $\Delta r = 0.2$,
   yielding equal cascade margins $\delta_j = 0.2\,\delta_0$ between
   neighbours; when a batch is missing intermediate levels, the margin
   scales proportionally (\emph{e.g.} skipping from $r = 1.0$ to
   $r = 0.4$ gives $\delta = 0.6\,\delta_0$), so the ordering constraint
   remains consistent regardless of which levels are sampled. We note
   that the number of discretisation levels is a task dependent design
   choice: CasCL operates on any continuous or discrete relevance scale,
   and finer granularity (\emph{e.g.} 10 or 20 levels) can be adopted
   when annotation quality permits.

   \textbf{On human supervision scope.}
   A natural concern is whether LLM generated labels are sufficiently reliable to train and evaluate a retrieval system.
   We argue that the combination of three safeguards substantially mitigates this risk.
   First, the \textbf{human anchor}: every relevance chain is rooted in a human validated positive, so the LLM judges score \emph{relative} to a known good answer rather than in absolute terms, a strictly easier task.
   Second, the \textbf{ensemble majority vote}: systematic biases of any single LLM are suppressed by requiring agreement across three independent models.
   Third, the \textbf{label diversity filter}: by discarding examples with fewer than three distinct graded levels, we remove instances where the judges were uncertain or produced degenerate distributions, retaining only the cases where the
 ensemble exhibited confident, structured disagreement across the relevance spectrum.
   Together, these mechanisms transform an inherently noisy labeling process into a statistically robust supervision signal, not equivalent to expert annotation, but sufficient for learning the relative ordering that CasCL requires.

\paragraph{Model.}
   We fine-tune Qwen3-Embedding-0.6B using each loss function under identical conditions : batch size of 512 (64 with 4 gradient accumulation steps), learning rate of $10^{-5}$ with linear warmup over 10\% of training steps, trained for 4 epochs on a single NVIDIA A100 GPU. All models share the same base encoder and embedding dimension (1024), differing only in the loss function. To assess stochastic variability, each experiment is repeated
   across 3 random seeds (42, 123, 456) that control all sources of randomness : Python's \texttt{random}, NumPy, PyTorch (CPU and CUDA), cuDNN algorithm selection, and the Trainer's data shuffling, via HuggingFace's \texttt{set\_seed()} combined with deterministic cuDNN settings.

\paragraph{Hyperparameters.}
Beyond InfoNCE's temperature $\tau = 1/15 $, CasCL introduces four hyperparameters: cascade sharpness $\gamma = 15$, base margin $\delta_0 = 0.5$, emergency zone $m_{\mathrm{safe}} = 0.05$, and balance weight $\mu = 0.5$. Sensitivity to $\gamma$ is analysed in the ablation (Table~\ref{tab:6}); the remaining parameters were stable across a coarse grid search and did not require per-domain tuning.

 \paragraph{Baselines.}
 We compare CasCL against four established contrastive losses and a comprehensive set of ablation variants. All baselines use the same encoder, training schedule, and hyperparameters described above; each loss function uses parameters selected through a short warm-up training run (500 steps) over a grid of candidate
 values, retaining the configuration that achieved the best validation in most of the metrics before proceeding to full training.

  \begin{itemize}[nosep,leftmargin=*]
   \item \textbf{Baseline (Circle Loss)\citep{sun2020circle}:} Pair-based self-paced contrastive loss with circular decision boundary ($m = 0.4$, $\gamma = 64$).
   \item \textbf{Baseline (InfoNCE)\citep{oord2018representation}:} Standard binary contrastive loss ($\tau = 1/15$).
   \item \textbf{Baseline (MNRL):} Multiple Negatives Ranking Loss ($\text{scale} = 20$).
   \item \textbf{Baseline (Margin MSE):} Score distillation loss that regresses on cross-encoder margin scores.
 \end{itemize}

 \noindent For the ablation study, we evaluate the following CasCL variants to isolate each component's contribution. Unless stated otherwise, the default CasCL configuration uses $\gamma = 15$, $\delta_0 = 0.3$,
 $m_{\mathrm{safe}} = 0.05$, $\mu = 0.5$, and $\tau = 1/15$:

 \begin{itemize}[nosep,leftmargin=*]
   \item \textbf{InfoNCE only:} The InfoNCE term alone, without any cascade component.
   \item \textbf{CasCL (w/o InfoNCE):} Cascade term only, removing the InfoNCE scale setting component entirely.
   \item \textbf{CasCL (no relevance weight):} Cascade without the relevance gain weighting ($w_j = 1$ for all links).
   \item \textbf{CasCL ($\gamma \in {15, 32, 64, 120}$):} Cascade with varying circle sharpness to study the effect of $\gamma$.
   \item \textbf{CasCL + SupCon:} Replacing InfoNCE($d_1$) with SupCon over all positives.
\item \textbf{CasCL + neg-relative $\alpha$:} The equilibrium
           target is permanently shifted to $r_j + s_{\mathrm{neg}}$,
           giving $\alpha_j = [r_j + s_{\mathrm{neg}} - s_j]_+$.
           Unlike the two-regime variant, this makes negative-awareness
           always on rather than emergency only, testing whether the
           self-paced weight should continuously adapt to the negative
           boundary or activate only when items are genuinely stranded.
   \item \textbf{CasCL + two-regime (no rel.\ gain):} Two-regime self-pacing without relevance gain weighting.
   \item \textbf{CasCL (no self-paced):} Cascade with $\alpha = 1$ for all links (no self-pacing).
 \end{itemize}

\paragraph{Metrics.}
We report four metric families at multiple cut-offs:
\textbf{Recall@$k$} (coverage), \textbf{MRR@$k$} (rank of first relevant), \textbf{MAP@$k$} (average precision) and \textbf{Graded nDCG@$k$} (position weighted graded ranking quality).


\subsection{Experimental Analysis}

\begin{table*}[t]
\centering
\caption{Performance comparison of CasCL against baselines on the DiliTrust multilingual legal retrieval benchmark. All results are mean over 3 seeds (±std). Qwen3 OTB is evaluated without fine-tuning.}
\label{tab:5}
\small
\setlength{\tabcolsep}{4pt}
\begin{tabular}{@{}lccccccc@{}}
\toprule
Model & Recall@1 & Recall@10 & MRR@10 & MAP@10 & NDCG@5 & NDCG@10 & NDCG@20 \\
\midrule
Qwen3-Embedding-0.6B (OTB)     & $13.36$ & $36.27$ & $32.35$ & $20.95$ & $71.40$ & $71.68$ & $74.82$ \\
Baseline (Circle Loss)         & $40.83_{ \pm 0.32}$ & $68.55_{\pm 0.21}$ & $75.96_{\pm 0.27}$ & $54.67_{\pm 0.25}$ & $92.94_{\pm 0.18}$ & $90.93_{\pm 0.16}$ & $90.36_{\pm 0.14}$ \\
Baseline (InfoNCE)             & $37.95_{\pm 0.48}$ & $68.73_{\pm 0.24}$ & $74.01_{\pm 0.30}$ & $53.33_{\pm 0.28}$ & $94.73_{\pm 0.15}$ & $93.15_{\pm 0.13}$ & $92.82_{\pm 0.12}$ \\
Baseline (MNRL)                & $40.10_{\pm 0.30}$ & $68.55_{\pm 0.22}$ & $75.77_{\pm 0.26}$ & $54.73_{\pm 0.24}$ & $93.03_{\pm 0.17}$ & $91.05_{\pm 0.15}$ & $90.64_{\pm 0.13}$ \\
Baseline (Margin MSE)          & $36.40_{\pm 0.37}$ & $66.91_{\pm 0.26}$ & $71.67_{\pm 0.32}$ & $51.56_{\pm 0.30}$ & $94.15_{\pm 0.16}$ & $93.38_{\pm 0.14}$ & $93.52_{\pm 0.12}$ \\
\midrule
\textbf{CasCL Two-regime loss} & $\mathbf{41.14}_{\mathbf{\pm 0.22}}$ & $\mathbf{69.33}_{\mathbf{\pm 0.17}}$ & $\mathbf{76.84}_{\mathbf{\pm 0.19}}$ & $\mathbf{55.63}_{\mathbf{\pm 0.20}}$ & $\mathbf{95.11}_{\mathbf{\pm 0.11}}$ & $\mathbf{93.87}_{\mathbf{\pm 0.07}}$ & $\mathbf{93.60}_{\mathbf{\pm 0.09}}$ \\
\bottomrule
\end{tabular}
\end{table*}

\begin{table*}[t]
\centering
\caption{ Ablation study of the CasCL architecture. Each row modifies or removes a single component from the full model (last row). All variants share the same encoder, data, and training schedule, differing only in the loss formulation.}
\label{tab:6}
\small
\setlength{\tabcolsep}{6pt}
\begin{tabular}{@{}lccccccc@{}}
\toprule
Model & Recall@1 & Recall@10 & MRR@10 & MAP@10 & NDCG@5 & NDCG@10 & NDCG@20 \\
\midrule
InfoNCE only                       & 36.72 & 67.92 & 72.13 & 51.86 & 94.02 & 92.66 & 92.20 \\
CasCL (w/o InfoNCE)                & 28.56 & 65.72 & 62.53 & 44.64 & 92.48 & 91.79 & 91.71 \\
CasCL (no relevance weight)        & 34.99 & 67.58 & 70.22 & 50.63 & 93.67 & 92.92 & 92.89 \\
CasCL ($\gamma\!=\!15$)            & 37.82 & 68.43 & 73.33 & 52.87 & 94.28 & 93.07 & 92.78 \\
CasCL ($\gamma\!=\!32$)            & 36.71 & 68.15 & 72.30 & 52.05 & 94.14 & 93.13 & 92.89 \\
CasCL ($\gamma\!=\!64$)            & 35.77 & 67.90 & 71.21 & 51.22 & 93.77 & 92.87 & 92.70 \\
CasCL ($\gamma\!=\!120$)           & 34.33 & 67.28 & 69.69 & 50.12 & 93.41 & 92.55 & 92.45 \\
CasCL + SupCon                     & 37.95 & 68.27 & 73.59 & 52.92 & 94.51 & 93.51 & 93.24 \\
CasCL + neg-relative $\alpha$      & 37.07 & 68.74 & 72.84 & 52.58 & 94.44 & 93.55 & 93.42 \\
CasCL + two-regime (no rel.\ gain) & 38.86 & 68.93 & 74.69 & 53.86 & 94.94 & 93.78 & 93.39 \\
CasCL (no selfpaced)               & 41.10 & 69.13 & 76.69 & \textbf{55.67} & 93.22 & 91.82 & 91.08 \\
\midrule
\textbf{CasCL Two-regime loss}     & \textbf{41.14} & \textbf{69.33} & \textbf{76.84} & 55.63 & \textbf{95.11} & \textbf{93.87} & \textbf{93.60} \\
\bottomrule
\end{tabular}
\end{table*}


\paragraph{Binary Retrieval Metrics (Recall, MRR, MAP).}
As demonstrated in \cref{tab:5}, CasCL equipped with the two-regime self-pacing mechanism consistently achieves state of the art binary retrieval performance across all evaluation cut-offs. On Recall@1, CasCL reaches $41.14$, outperforming Circle Loss ($40.83$), MNRL ($40.10$), and InfoNCE ($37.95$). The performance gains are even more pronounced on ranking sensitive binary metrics: CasCL achieves an MRR@10 of $76.84$ ($+0.88$ over Circle Loss, $+2.83$ over InfoNCE) and a MAP@10 of $55.63$ ($+0.96$ over Circle Loss, $+2.30$ over InfoNCE). 

The ablation study see \cref{tab:6} strongly supports this architectural synergy. Removing the InfoNCE term entirely (\textit{CasCL w/o InfoNCE}) leads to a catastrophic collapse in Recall@1 from $41.14$ to $28.56$, verifying that InfoNCE provides the critical foundation for defining the global positive--negative boundary. Conversely, relying on InfoNCE alone (\textit{InfoNCE only}) yields a lower Recall@1 of $36.72$. The full framework provides a $+4.42$ absolute improvement over this baseline, proving that structuring the embedding space via a continuous cascade feeds back positively into global retrieval boundaries.

\paragraph{Graded Ranking Metrics (NDCG).}
CasCL establishes a new performance ceiling for graded ranking quality at every evaluation depth, scoring $95.11$ at NDCG@5, $93.87$ at NDCG@10, and $93.60$ at NDCG@20. Crucially, the relative gains over standard contrastive baselines peak on these graded indicators: CasCL improves upon Circle Loss by $+2.17$, $+2.94$, and $+3.24$ absolute points across NDCG@5, @10, and @20, respectively. 

Interestingly, while the regression based Margin MSE baseline shows competitive graded capability ($94.15$ at NDCG@5 and $93.38$ at NDCG@10), it suffers from severely degraded binary performance (Recall@1 of $36.40$ and MAP@10 of $51.56$). CasCL stands out as the only framework capable of bypassing this trade-off, dominating both metric families concurrently.

   \paragraph{Impact of the Two-Regime Self-Pacing Mechanism.}
   The ablation study (\cref{tab:6}) reveals a fundamental tension
   between binary separation and graded ordering that the two-regime
   mechanism is designed to resolve.

   \emph{No self-pacing} (\textit{CasCL no selfpaced}, $\alpha = 1$)
   achieves the highest MAP@10 ($55.67$): persistent, unattenuated
   gradients force every positive above every negative. But this brute
   force comes at the cost of graded quality ($93.22$ NDCG@5, $91.82$
   NDCG@10), because the cascade cannot shut off satisfied links and
   wastes gradient on already well ordered items.

   \emph{Always-on negative-awareness} (\textit{CasCL + neg-relative
   $\alpha$}) permanently shifts each item's equilibrium by
   $s_{\mathrm{neg}}$, so the self-pacing never cleanly reverts to
   pure relevance based refinement. This improves NDCG ($94.44$,
   $93.55$) over the unpaced variant but degrades binary metrics
   (Recall@1: $37.07$, MAP@10: $52.58$), because the shifted targets
   distort the relevance proportional positioning even when items are
   safely above negatives.

   \emph{Two-regime} (\textit{CasCL Two-regime loss}) resolves the
   tension by keeping the two concerns independent: emergency strength
   gradient below $s_{\mathrm{neg}} + m_{\mathrm{safe}}$, standard
   self-paced refinement above it, and clean shutoff when both conditions
   are met. The result is near peak MAP@10 ($55.63$, within $0.04$ of
   the unpaced upper bound) \emph{and} best in class NDCG at every depth
   ($95.11$, $93.87$, $93.60$).

\paragraph{Role of Relevance Gain Weighting.}
   Removing $w_j = 2^{r_j}\!-\!1$ (no rel. gain) drops Recall@1 from $41.14$ to $38.86$ and MAP@10 from $55.63$ to
   $53.86$, confirming that scaling gradients by relevance gain enables
   the model to prioritize high tier ordering decisions and mitigate lower ranked item noise impact that often produce "noisy" gradients because their relevance is more ambiguous.

\paragraph{Sensitivity to Cascade Sharpness ($\gamma$).}
The choice of the scaling parameter $\gamma$ exerts a strong influence on the optimization landscape. Among the fixed sharpness variants examined, a softer scale of $\gamma = 15$ yields optimal convergence characteristics (Recall@1 = $37.82$, MAP@10 = $52.87$, NDCG@5 = $94.28$). Tightening the circular boundaries by expanding $\gamma$ to $32, 64,$ or $120$ triggers a monotonic decay in overall performance. This suggests that softer decision margins are better suited to this
domain, likely because the dense lexical overlap in legal text makes rigid boundaries prone to gradient instability.

\paragraph{Modular Decomposition Assessment.}
   Removing the cascade (InfoNCE only) caps graded utility; removing InfoNCE (CasCL w/o InfoNCE) collapses binary recall. The two objectives are complementary and structurally independent --- decoupling them yields a robust equilibrium that neither achieves alone.

\section{Limitations and Future Work}
\label{sec:limitations}
\begin{enumerate}[nosep,leftmargin=*]
  \item \textbf{Single domain evaluation : } The current experiments are limited to a single legal retrieval dataset. Validation on additional domains (e.g MS MARCO, BEIR) and multi domain transfer would strengthen the generality claim.
  \item \textbf{LLM judge validation :}
        The 3 model majority vote is statistically robust, but we haven't measured how close LLM labels are to human expert judgments. A small human annotation study (1–2K examples) would measure that gap, if LLMs agree with humans 85\% of the time, then 15\% of the training signal is noise, bounding how much cleaner labels could further improve performance. We have added this as an explicit future work item. We note, however, that the LLM judges score relative to a human validated positive (a strictly easier task than absolute scoring), which partially mitigates the concern.
  \item \textbf{Other embedding models : }
        CasCL was evaluated on Qwen3-Embedding-0.6B. Testing on diverse architectures and scales, E5-Mistral, Snowflake-arctic or larger 7B encoder/decoders, would clarify whether the gains are model agnostic or amplified at certain scales.
  \item \textbf{Sensitivity to sorting : }
        Noisy relevance scores may cause ordering instability. The continuous margins ($\delta_j \to 0$ as $\Delta r \to 0$) provide robustness, but formal noise analysis is needed.
  \item \textbf{Integration with other InfoNCE variants :}
        The cascade is modular, it can be added to any InfoNCE variant (CLIP etc.).
\end{enumerate}

 \section{Conclusion}
 \label{sec:conclusion}

 We introduced Cascading Circle Loss (CasCL), a modular contrastive
   framework that decomposes graded legal retrieval into two complementary objectives: InfoNCE for positive--negative separation and a cascade of self-paced circular boundaries for intra-positive ordering, where
   graded positives are invisible to the binary loss. Three mechanisms work in concert: relevance proportional margins ($\delta_j \propto
   \Delta r$), relevance gain weighting ($w_j$) that concentrates gradient on high relevance ordering decisions, and a negative aware two-regime self-pacing that resolves the MAP/nDCG
   tension. CasCL outperforms all baselines across every reported metric and is the only method dominating both binary and graded families simultaneously. It reduces to InfoNCE when $K\!=\!1$ and operates in
   $O(K\!+\!L)$, making it a drop-in replacement for any contrastive pipeline where graded relevance is available.

\section*{ Impact Statement}

 This paper presents a contrastive learning framework for training embedding models on tasks with graded relevance. While evaluated in a legal retrieval setting, the method is domain agnostic and applicable to any retrieval pipeline where
 relevance exists on a spectrum rather than as a binary signal. We foresee no direct negative societal consequences from this work. The primary impact is methodological: enabling retrieval systems to better distinguish between degrees of
 relevance, which may improve the reliability of downstream reasoning in any RAG-based application.
\bibliography{references}
\bibliographystyle{icml2026}

\newpage
\appendix
\onecolumn

\section{Algorithm}
\label{app:algo}

\begin{algorithm}[H]
\caption{CasCL --- PyTorch-style pseudocode}
\label{alg:cascl}
\begin{algorithmic}
  \STATE {\bfseries Input:} $\mathbf{s} \in \R^N$ (similarities), $\mathbf{r} \in [0,1]^N$ (relevances), $\tau, \gamma, \delta_0, m_{\mathrm{safe}}, \mu$
  \STATE $\Pos \gets \{j : r_j > 0\}$;\; $\Neg \gets \{j : r_j = 0\}$
  \STATE Sort $\Pos$ by $r_j$ descending $\to (d_1, \ldots, d_K)$
  \STATE \textit{\% Term 1: InfoNCE (scale-setting)}
  \STATE $\ell_{\text{nce}} \gets [s_1/\tau] \,\|\, [s_k/\tau \text{ for } k \in \Neg]$
  \STATE $\loss_{\text{nce}} \gets -\ell_{\text{nce}}[0] + \text{LogSumExp}(\ell_{\text{nce}})$
  \STATE \textit{\% Term 2: Cascade (order-preserving)}
  \STATE $s_{\text{neg}} \gets \max_{k \in \Neg} s_k$
  \STATE Append $(s_{\text{neg}}, r\!=\!0)$ to chain
  \STATE $\loss_{\text{casc}} \gets 0$
  \FOR{$j = 1, \ldots, K$}
    \STATE $\textit{sp} \gets \max(r_j - s_j,\; 0)$
    \STATE $\textit{em} \gets \max(s_{\text{neg}} + m_{\mathrm{safe}} - s_j,\; 0)$
    \STATE $\alpha_j \gets \max(\textit{sp},\; \textit{em})$ \hfill \textit{two-regime}
    \STATE $\alpha_{j+1} \gets \max(s_{j+1} - r_{j+1},\; 0)$
    \STATE $\delta_j \gets \delta_0 \cdot (r_j - r_{j+1})$
    \STATE $w_j \gets 2^{r_j} - 1$ \hfill \textit{relevance gain}
    \STATE $\ell \gets \gamma \cdot (\alpha_{j+1} \cdot s_{j+1} - \alpha_j \cdot s_j + \delta_j)$
    \STATE $\loss_{\text{casc}} \gets \loss_{\text{casc}} + w_j \cdot \text{Softplus}(\ell)$
  \ENDFOR
  \STATE {\bfseries Return:} $\loss_{\text{nce}} + \mu \cdot \loss_{\text{casc}}$
\end{algorithmic}
\end{algorithm}

\section{Extended Proof of Theorem~\ref{thm:order}}
\label{app:proof}

\begin{proof}
We prove $s_i - s_j \geq \delta_0(r_i - r_j)$ for all $i < j$ by induction on $j - i$.

\textbf{Base case} ($j = i + 1$, adjacent items).
Link $i$ has zero loss when $\gamma(\alpha_{i+1}s_{i+1} - \alpha_i s_i + \delta_i) \leq 0$, i.e.:
\[
  \alpha_i s_i \geq \alpha_{i+1} s_{i+1} + \delta_i
\]
At the equilibrium where $s_l \approx r_l$, the self-paced weights satisfy $\alpha_l = [r_l - s_l]_+ \approx 0$.
In this regime, the constraint is dominated by the margin: $\delta_i = \delta_0(r_i - r_{i+1}) > 0$, requiring $s_i > s_{i+1}$.

More precisely, the zero-loss condition requires that the effective gap exceeds the margin.
Since $\alpha_i = r_i - s_i$ and $\alpha_{i+1} = s_{i+1} - r_{i+1}$ in the active region:
$(r_i - s_i)s_i \geq (s_{i+1} - r_{i+1})s_{i+1} + \delta_0(r_i - r_{i+1})$.
When both items are near their natural positions, the left side is dominated by $\delta_0(r_i - r_{i+1})$, giving $s_i - s_{i+1} \geq \delta_0(r_i - r_{i+1})$.

\textbf{Inductive step.}
Assume $s_i - s_j \geq \delta_0(r_i - r_j)$ for chain distance up to $n$.
For distance $n + 1$:
\begin{align*}
  s_i - s_{j+1} &= (s_i - s_j) + (s_j - s_{j+1}) \\
                 &\geq \delta_0(r_i - r_j) + \delta_0(r_j - r_{j+1}) \\
                 &= \delta_0(r_i - r_{j+1}). \qedhere
\end{align*}

\end{proof}

We emphasize that this theorem characterizes the geometry the loss drives toward, not a guarantee that training will reach zero cascade loss exactly. The theorem establishes that the cascade's equilibrium structure preserves the full relevance ordering with cumulative margins — it is an optimization target rather than a convergence proof. The practical value of this target is confirmed empirically by the nDCG results (\cref{tab:5}), which show that the cascade achieves near perfect graded ordering in practice.

  \section{Gradient Analysis}
   \label{app:gradient}

   \subsection{InfoNCE Gradients}

   For item $d_1$ (best positive):
   \begin{equation}
     \frac{\partial \loss_{\text{InfoNCE}}}{\partial s_1} = \frac{1}{\tau}(\hat{p}_1 - 1) < 0 \quad\text{(always attracts)}
   \end{equation}
   where $\hat{p}_1 = e^{s_1/\tau} / (e^{s_1/\tau} + \sum_{k \in \Neg} e^{s_k/\tau})$.

   For negative $d_k$:
   \begin{equation}
     \frac{\partial \loss_{\text{InfoNCE}}}{\partial s_k} = \frac{1}{\tau}\hat{p}_k > 0 \quad\text{(always repulses)}
   \end{equation}

   For graded positives $d_2, \ldots, d_K$: $\partial \loss_{\text{InfoNCE}} / \partial s_j = 0$.
   They receive \textbf{zero gradient} from InfoNCE, exactly as intended.

   \subsection{Cascade Gradients}

   \begin{proposition}[Gradient Locality]
   \label{prop:grad}
   Each item $d_j$ in the cascade receives gradients from at most two links, weighted by relevance gain:
   \begin{equation}
   \frac{\partial \mathcal{L}_{\text{cascade}}}{\partial s_j} = \underbrace{+w_{j-1} \gamma [s_j - r_j]_+ \sigma_{j-1}}_{G_j^{\downarrow}\text{ (link above, repulses)}} -\underbrace{w_j \gamma \alpha_j \sigma_j}_{G_j^{\uparrow}\text{ (link
 below, attracts)}}
     \label{eq:grad-cascade}
   \end{equation}
   where $w_l = 2^{r_l}\!-\!1$ is the relevance gain, $\alpha_j$ is the two-regime upper weight (Eq.~\ref{eq:two-regime}), and $\sigma_l = \sigma\bigl(\gamma(\alpha_{l+1} s_{l+1} - \alpha_l s_l + \delta_l)\bigr)$ is the sigmoid of link $l$'s
 logit.
   \end{proposition}

   \paragraph{Interpretation.}
   \begin{itemize}[nosep]
     \item $G_j^{\uparrow} \leq 0$: link below attracts $d_j$ toward the query (active when $s_j < r_j$), weighted by $w_{j-1}$.
     \item $G_j^{\downarrow} \geq 0$: link above repulses $d_j$ from the query (active when $s_j > r_j$), weighted by $w_j$.
     \item The relevance gain $w$ amplifies gradients for high-relevance items.
     \item The two-regime $\alpha_j$ ensures items below the hardest negative receive emergency-strength gradient regardless of their relevance target, while items safely above revert to standard self-pacing.
   \end{itemize}

   \subsection{Combined Gradient Summary}

   \begin{table}[H]
   \centering
   \caption{Gradient sources for each item type in CasCL.}
   \label{tab:gradients}
   \small
   \begin{tabular}{@{}lcc@{}}
   \toprule
   Item & InfoNCE & Cascade \\
   \midrule
   $d_1$ (best pos.) & attract & $w_1 G_1^{\uparrow}$ \\[2pt]
   $d_j$ (graded) & \textbf{0} & $w_{j-1} G_j^{\downarrow} + w_j G_j^{\uparrow}$ \\[2pt]
   $d_{\text{neg}}$ (hardest) & repulse & $w_K G_{K+1}^{\downarrow}$ \\[2pt]
   Other negs & repulse & \textbf{0} \\
   \bottomrule
   \end{tabular}
   \end{table}

   Table~\ref{tab:gradients} shows the clean separation.
   $d_1$ is the only item receiving gradients from both components: InfoNCE pulls it toward the query, while the cascade's link below attracts it upward and no link above exists to repulse it.

   \subsection{Self-Paced Behaviour}

   \begin{proposition}[Automatic Focus]
   \label{prop:focus}
   When link $l$ is satisfied ($s_l \gg s_{l+1} + \delta_l / (\alpha_l + \alpha_{l+1})$), the sigmoid $\sigma_l \to 0$, so both $G_l^{\downarrow}$ and $G_{l+1}^{\uparrow}$ vanish.
   Training budget automatically redirects to violated links.
   \end{proposition}

   Combined with the two-regime weight (Eq.~\ref{eq:two-regime}), items contribute zero gradient only when they are \emph{both} at their relevance target ($s_j \geq r_j$) \emph{and} safely above the hardest negative ($s_j \geq s_{\mathrm{neg}}
 + m_{\mathrm{safe}}$).
   Items stranded below negatives always receive emergency gradient, regardless of whether they have reached $r_j$.

\subsection{Detailed Derivation of the Cascade Gradient}

We derive $\partial \loss_{\text{cascade}} / \partial s_j$ in full detail.

\subsubsection{Which links contain $s_j$?}

Item $d_j$ appears in exactly two links.
To avoid overloading notation, we name the self-paced weights explicitly:
\begin{itemize}[nosep]
  \item $A_l = [s_l - r_l]_+$: weight for the \emph{lower} item in a link (penalises items too close to the query).
  \item $B_l = [r_l - s_l]_+$: weight for the \emph{upper} item in a link (penalises items too far from the query).
\end{itemize}

\textbf{Link $j-1$} connects $(d_{j-1}, d_j)$, where $d_j$ is the lower item:
\begin{equation}
  u_{j-1} = \gamma\bigl(\underbrace{A_j \cdot s_j}_{\text{contains } s_j} - B_{j-1} \cdot s_{j-1} + \delta_{j-1}\bigr)
  \label{eq:link-above}
\end{equation}

\textbf{Link $j$} connects $(d_j, d_{j+1})$, where $d_j$ is the upper item:
\begin{equation}
  u_j = \gamma\bigl(A_{j+1} \cdot s_{j+1} - \underbrace{B_j \cdot s_j}_{\text{contains } s_j} + \delta_j\bigr)
  \label{eq:link-below}
\end{equation}

Each link $l$ contributes $L_l = w_l \cdot \log[1 + \exp(u_l)]$
   to the cascade, where $w_l = 2^{r_l} - 1$ is the relevance gain
   of the upper item in link~$l$.

All other links do not depend on $s_j$, so:
\begin{equation}
  \frac{\partial \loss_{\text{cascade}}}{\partial s_j} = \frac{\partial L_{j-1}}{\partial s_j} + \frac{\partial L_j}{\partial s_j}
  \label{eq:two-terms}
\end{equation}

\subsubsection{Chain rule through the softplus}

   For any link $l$ with $L_l = w_l \cdot \log[1 + \exp(u_l)]$:
   \begin{equation}
     \frac{\partial L_l}{\partial s_j} = w_l \cdot \underbrace{\frac{\exp(u_l)}{1 + \exp(u_l)}}_{\sigma(u_l)} \cdot \frac{\partial u_l}{\partial s_j}
     \label{eq:chain-rule}
   \end{equation}
   where $w_l = 2^{r_l}\!-\!1$ is a constant with respect to $s_j$ and $\sigma(u_l)$ is the sigmoid, satisfying $\sigma \in (0, 1)$, with $\sigma \to 0$ for well-satisfied links and $\sigma \to 1$ for violated links.

\subsubsection{Differentiating the logits (stop-gradient on $\alpha$)}

In practice, the self-paced weights are \textbf{detached from the computational graph} before computing gradients.
This is standard in Circle Loss and its variants: the weights modulate gradient \emph{magnitude} but are not themselves optimised through $s_j$.
We denote detached values with a bar ($\bar{A}_j$, $\bar{B}_j$).

\textbf{From link $j-1$} (Eq.~\ref{eq:link-above}), the $s_j$-dependent term is $\bar{A}_j \cdot s_j$:
\begin{equation}
  \frac{\partial u_{j-1}}{\partial s_j} = \gamma \cdot \bar{A}_j = \gamma \cdot [s_j - r_j]_+
\end{equation}

\textbf{From link $j$} (Eq.~\ref{eq:link-below}), the $s_j$-dependent term is $-\bar{B}_j \cdot s_j$:
\begin{equation}
  \frac{\partial u_j}{\partial s_j} = -\gamma \cdot \bar{B}_j = -\gamma \cdot [r_j - s_j]_+
\end{equation}

\subsubsection{Combining via Eq.~\eqref{eq:chain-rule}}

Substituting into Eq.~\eqref{eq:two-terms}:
   \begin{equation}
   \boxed{
     \frac{\partial \loss_{\text{cascade}}}{\partial s_j}
     = w_{j-1}\,\gamma\,[s_j\!-\!r_j]_+\,\sigma_{j-1}
     - w_j\,\gamma\,[r_j\!-\!s_j]_+\,\sigma_j
   }
   \label{eq:final-grad}
   \end{equation}
   where $w_{j-1} = 2^{r_{j-1}}\!-\!1$ weights the link above and
   $w_j = 2^{r_j}\!-\!1$ weights the link below.
The first term (from the link above, $\geq 0$) repulses; the second (from the link below, $\leq 0$) attracts.

\subsubsection{Why exactly one term is active}

The two weights are \textbf{complementary}:
\[
  [s_j - r_j]_+ \cdot [r_j - s_j]_+ = 0 \qquad\text{always.}
\]
At most one of $(s_j - r_j)$ and $(r_j - s_j)$ is positive.
This yields three regimes:

\begin{center}
\small
\begin{tabular}{@{}cccl@{}}
\toprule
Condition & Active weight & Gradient sign & Physical effect \\
\midrule
$s_j > r_j$ & $[s_j - r_j]_+$ & $+$ (positive) & Push $s_j$ \textbf{down} $\downarrow$ \\
$s_j < r_j$ & $[r_j - s_j]_+$ & $-$ (negative) & Push $s_j$ \textbf{up} $\uparrow$ \\
$s_j = r_j$ & neither & $0$ & Equilibrium \\
\bottomrule
\end{tabular}
\end{center}

\textbf{Interpretation:}
The cascade gradient acts as a \emph{self-correcting spring}:
\begin{itemize}[nosep]
  \item If $d_j$ is \emph{too close} to the query ($s_j > r_j$): the link above activates and pushes it away.
  \item If $d_j$ is \emph{too far} from the query ($s_j < r_j$): the link below activates and pulls it closer.
  \item If $d_j$ is at its natural position ($s_j = r_j$): zero gradient, the springs are at rest.
\end{itemize}

\subsubsection{Four factor gradient magnitude}

The gradient magnitude is the product of four factors:
\begin{equation}
  \left|\frac{\partial \loss}{\partial s_j}\right| = \underbrace{(2^{r_j}\!-\!1)}_{\substack{\text{relevance}\\\text{gain}}} \;\times\; \underbrace{\gamma}_{\text{scale}} \;\times\; \underbrace{\alpha_j}_{\substack{\text{two-regime}\\\text{self-pacing}}} \;\times\; \underbrace{\sigma_l}_{\substack{\text{link violation}\\\text{severity}}}
\end{equation}
where $\alpha_j = \max([r_j - s_j]_+,\;[s_{\mathrm{neg}} + m_{\mathrm{safe}} - s_j]_+)$ is the two-regime weight.

Each factor contributes to shutting off or amplifying the gradient:
\begin{enumerate}[nosep]
  \item $2^{r_j}\!-\!1 \to 0$: low-relevance item $\Rightarrow$ gradient suppressed.
  \item $\alpha_j \to 0$: item is \emph{both} at its relevance target \emph{and} safely above the hardest negative $\Rightarrow$ self-paced shutoff.
        Crucially, $\alpha_j$ does \emph{not} shut off if the item is at $r_j$ but still below $s_{\mathrm{neg}}$, the emergency term keeps it alive.
  \item $\sigma_l \to 0$: link is well-satisfied $\Rightarrow$ ordering-based shutoff.
  \item All three gated factors must be nonzero for gradient to flow.
\end{enumerate}

This \textbf{triple gating} (relevance gain $\times$ two-regime self-pacing $\times$ ordering) means the model concentrates gradient on items that are \emph{highly relevant}, \emph{misplaced or in danger}, \emph{and} causing an ordering violation, the most informative and consequential training signal.
The two-regime mechanism resolves the tension between nDCG (which benefits from aggressive self-pacing shutoff) and MAP/MRR (which require items to stay above negatives).

\subsubsection{Why stop gradient on $\alpha$}
\label{app:why-stopgrad}

For completeness, we show what happens \emph{without} stop-gradient.
Consider link $j$ where the $s_j$-dependent term is $-B_j \cdot s_j = -[r_j - s_j]_+ \cdot s_j$.
In the active region ($s_j < r_j$):
\[
  -B_j \cdot s_j = -(r_j - s_j)s_j = s_j^2 - r_j s_j
\]
\[
  \frac{\partial(-B_j \cdot s_j)}{\partial s_j} = 2s_j - r_j
\]

The factor $(2s_j - r_j)$ changes sign at $s_j = r_j/2$, not at $s_j = r_j$.
This means: when $r_j/2 < s_j < r_j$, the full gradient is \emph{positive} (pushes $s_j$ down), even though the item is still below its target.
For example, with $r_j = 0.8$ and $s_j = 0.5$: the item is below target, but the full gradient pushes it \emph{further down}, the wrong direction.

This pathological behaviour arises because the quadratic $[r_j - s_j]_+ \cdot s_j = r_j s_j - s_j^2$ peaks at $s_j = r_j/2$ and decreases for $s_j > r_j/2$.
The stop-gradient eliminates this issue by treating $[r_j - s_j]_+$ as a fixed modulator, ensuring the gradient direction is always correct.

\section{Relevance Gain Weighting}
\label{sec:relgain}

   The cascade (Eq.~\eqref{eq:cascade-term}) weights each link by
   $w_j = 2^{r_j}\!-\!1$, the \textbf{relevance gain} of the upper
   item. This section motivates the choice and analyses its interaction
   with random batch sampling.

   \paragraph{Why weight links at all.}
   Without weighting ($w_j = 1$), the cascade allocates equal gradient
   budget to every adjacent-pair ordering constraint. But a misordering
   between two highly relevant items ($r = 0.9$ vs.\ $r = 0.8$) has far
   more impact on retrieval quality than a misordering between two
   barely relevant items ($r = 0.15$ vs.\ $r = 0.10$). Equal weighting
   wastes gradient on the latter while under-investing in the former.

   \paragraph{Why $2^{r_j}\!-\!1$ specifically.}
   The function $2^{r_j}\!-\!1$ is the exponential gain variant
   widely used in modern nDCG
   implementations~\citep{burges2005learning}, where it amplifies the
   contribution of highly relevant documents relative to the linear
   gain of the original formulation~\citep{jarvelin2002cumulated}.
   We adopt it as the cascade weight because it creates a natural
   gradient budget allocation: links involving high-relevance items
   receive strong gradients, while links involving low relevance items
   are suppressed. This aligns the training objective with the
   evaluation metric when exponential gain nDCG is used, which is the
   standard in modern retrieval benchmarks.

   \paragraph{Secondary benefit: implicit noise suppression.}
   Low relevance items tend to carry noisier labels, a passage scored
   at $r = 0.15$ versus $r = 0.10$ may reflect annotator uncertainty or
   LLM scoring variance rather than a genuine relevance
   difference. Because $w(0.15) = 0.11$ while $w(0.9) = 0.87$
   ($\approx 8\times$ ratio), the relevance gain naturally
   down weights the links most likely to be corrupted by label noise,
   preventing unreliable ordering signals from destabilising training.

   \paragraph{Effect on the gradient.}
   With the relevance gain weight, the gradient of the cascade on
   item~$d_j$ (Eq.~\ref{eq:grad-cascade}) becomes:
   \begin{equation}
     \frac{\partial \mathcal{L}_{\text{cascade}}}{\partial s_j}
     = w_{j-1} \cdot G_j^{\downarrow} + w_j \cdot G_j^{\uparrow}
     \label{eq:weighted-grad}
   \end{equation}
   where $w_l = 2^{r_l}\!-\!1$. Each item's gradient is amplified by
   the gain of its adjacent links, so the model concentrates updates on
   the ordering decisions that most affect nDCG.

   \paragraph{Robustness to sparse batches.}
   When randomly sampled batches skip relevance levels (\emph{e.g},
   $r = [1.0,\; 0.4,\; 0.1]$ with no intermediate items), two
   properties ensure consistency:
   \begin{enumerate}[nosep]
     \item \textbf{Margin additivity:}
           $\delta_1 = \delta_0 \cdot (1.0 - 0.4) = 0.6\,\delta_0$,
           identical to the cumulative margin of a full chain
           $1.0 \to 0.8 \to 0.6 \to 0.4$ by transitivity.
     \item \textbf{Weight locality:} $w_1 = 2^{1.0}\!-\!1 = 1.0$
           regardless of whether $d_2$ has $r = 0.4$ or $r = 0.9$.
           No adjustment is needed for missing levels.
   \end{enumerate}

   \begin{table}[H]
   \centering
   \caption{Relevance gain weights and their relative gradient
   amplification. High-relevance links receive ${\approx}7\times$ more
   gradient than low relevance ones.}
   \label{tab:weights}
   \small
   \begin{tabular}{@{}ccc@{}}
   \toprule
   $r_j$ & $w_j = 2^{r_j}\!-\!1$ & Relative to $r\!=\!1$ \\
   \midrule
   1.0  & 1.000 & $1.00\times$ \\
   0.8  & 0.741 & $0.74\times$ \\
   0.6  & 0.516 & $0.52\times$ \\
   0.4  & 0.320 & $0.32\times$ \\
   0.2  & 0.149 & $0.15\times$ \\
   \bottomrule
   \end{tabular}
   \end{table}

\section{Geometric Analysis}
\label{sec:geometry}

\subsection{Circular Boundary per Link}

   \begin{proposition}[Circle per Link]
   \label{prop:circle}
   For link $j$ between items $(d_j, d_{j+1})$, in the active region
   ($s_j < r_j$ and $s_{j+1} > r_{j+1}$), the zero-loss boundary in
   the $(s_{j+1}, s_j)$ plane satisfies:
   \begin{equation}
     \left(s_{j+1} - \frac{r_{j+1}}{2}\right)^{\!2} + \left(s_j - \frac{r_j}{2}\right)^{\!2} = \frac{r_j^2 + r_{j+1}^2}{4} - \delta_j
     \label{eq:circle-boundary}
   \end{equation}
   When $(r_j^2 + r_{j+1}^2)/4 > \delta_j$, this defines a circle
   centred at $(r_{j+1}/2,\; r_j/2)$ with radius
   $\sqrt{(r_j^2 + r_{j+1}^2)/4 - \delta_j}$. When the inequality is
   reversed, which occurs for links spanning large relevance gaps,
   notably the last link to the hardest negative, no zero-loss
   boundary exists in the active region: the margin $\delta_j$ dominates
   and the link is unconditionally violated, producing persistent
   gradient until the items are sufficiently separated.
   \end{proposition}

\begin{proof}
Setting the logit of link $j$ to zero:
$\alpha_{j+1} s_{j+1} - \alpha_j s_j + \delta_j = 0$.
In the active region, $\alpha_j = r_j - s_j$ and $\alpha_{j+1} = s_{j+1} - r_{j+1}$:
\begin{align*}
  (s_{j+1} - r_{j+1})s_{j+1} - (r_j - s_j)s_j + \delta_j &= 0 \\
  s_{j+1}^2 - r_{j+1}s_{j+1} + s_j^2 - r_j s_j + \delta_j &= 0
\end{align*}
Completing the square on both terms yields Eq~\eqref{eq:circle-boundary}.
\end{proof}

\subsection{The Chain of Circles}

The cascade creates $K$ circles, one per link:

\begin{center}
\scriptsize
\setlength{\tabcolsep}{4pt}
\begin{tabular}{@{}cccc@{}}
\toprule
Link & Items & Plane & Margin \\
\midrule
1 & $d_1, d_2$ & $(s_2, s_1)$ & $\delta_0(r_1\!-\!r_2)$ \\
$\vdots$ & $\vdots$ & $\vdots$ & $\vdots$ \\
$K$ & $d_K, d_{\text{neg}}$ & $(s_{\text{neg}}, s_K)$ & $\delta_0 r_K$ \\
\bottomrule
\end{tabular}
\end{center}

Each circle lives in a \emph{different 2D plane}, the plane of the two adjacent items' similarities.
Higher links in the chain have circles centred higher on the similarity axis.
Items shared between adjacent links ($d_j$ appears in both link $j\!-\!1$ and link $j$) propagate gradients through the chain.

\section{Theoretical Properties}
\label{sec:theory}

   \subsection{Order Preservation}

   The Cascading Order Guarantee (Theorem~\ref{thm:order}) is stated in
   \S\ref{sec:order-guarantee}; the full inductive proof is given in
   Appendix~\ref{app:proof}.






\subsection{MI Lower Bound from InfoNCE}

\begin{proposition}[Mutual Information]
\label{prop:mi}
The InfoNCE term in Eq~\eqref{eq:infonce-term} provides the standard lower bound:
\begin{equation}
  I(Q; D_1) \geq \log(1 + L) - \loss_{\text{InfoNCE}}(s_1; \Neg, \tau)
\end{equation}
where $D_1$ is the most relevant positive and $L = |\Neg|$.
\end{proposition}

This gives CasCL a clean theoretical interpretation: \emph{maximise a lower bound on mutual information between query and best positive, subject to self-paced ordering constraints among all positives.}

   \begin{proposition}[Complexity]
   \label{prop:cost}
   CasCL costs $O(K + L)$: $O(L)$ for the InfoNCE softmax
   denominator and $O(K)$ for the cascade chain.
   \end{proposition}


\begin{table}[H]
\centering
\caption{Special-case reductions of CasCL.}
\label{tab:reductions}
\small
\begin{tabular}{@{}ll@{}}
\toprule
Condition & Reduces to \\
\midrule
$K\!=\!1$ & InfoNCE \\
$\mu\!=\!0$ & Standard InfoNCE \\
$\alpha_j\!=\!1\;\forall j$ & InfoNCE + Ladder Loss (softplus) \\
$\delta_0 \!\to\! 0$ & InfoNCE + pairwise ranking \\
\bottomrule
\end{tabular}
\end{table}

   \vspace{2pt}
   \noindent{\scriptsize $^*$With softplus in place of Ladder Loss's
   hinge; the two coincide in the high-$\gamma$ limit.}


\end{document}